\documentclass[letterpaper]{article} 
\usepackage[submission]{aaai2026}  
\usepackage{times}  
\usepackage{helvet}  
\usepackage{courier}  
\usepackage[hyphens]{url}  
\usepackage{graphicx} 
\usepackage{natbib}  
\usepackage{caption} 
\usepackage{algorithm}
\usepackage{algorithmic}
\usepackage{multirow}
\usepackage{makecell}
\usepackage{ragged2e}
\usepackage{amssymb}
\usepackage{amsmath}
\usepackage{booktabs}
\usepackage{threeparttable}
\usepackage{newfloat}
\usepackage{listings}
\DeclareCaptionStyle{ruled}{labelfont=normalfont,labelsep=colon,strut=off} 
\floatstyle{ruled}
\newfloat{listing}{tb}{lst}{}
\floatname{listing}{Listing}
\title{UniAlignment: Semantic Alignment for Unified Image Generation,
Understanding, Manipulation and Perception}

\author{
    Xinyang Song\textsuperscript{\rm 1,2}\thanks{Work done during internship at Ant Group.}, Libin Wang\textsuperscript{\rm 3}\thanks{Project lead.},  Weining Wang\textsuperscript{\rm 2}, 
    Shaozhen Liu\textsuperscript{\rm 2},\\ Dandan Zheng\textsuperscript{\rm 3}, Jingdong Chen\textsuperscript{\rm 3}, Qi Li\textsuperscript{\rm 1,2}, Zhenan Sun\textsuperscript{\rm 1,2}
}
 \affiliations{
 \textsuperscript{\rm 1}School of Artificial Intelligence, University of Chinese Academy of Sciences\\
 \textsuperscript{\rm 2}Institute of Automation, Chinese Academy of Sciences\\
 \textsuperscript{\rm 3}AntGroup\\

 }

\begin{document}

\maketitle

\begin{abstract}

The remarkable success of diffusion models in text-to-image generation has sparked growing interest in expanding their capabilities to a variety of multi-modal tasks, including image understanding, manipulation, and perception. 
These tasks require advanced semantic comprehension across both visual and textual modalities, especially in scenarios involving complex semantic instructions.
However, existing approaches often rely heavily on vision-language models (VLMs) or modular designs for semantic guidance, leading to fragmented architectures and computational inefficiency. 
To address these challenges, we propose UniAlignment, a unified multimodal generation framework within a single diffusion transformer. 
UniAlignment introduces a dual-stream diffusion training strategy that incorporates both intrinsic-modal semantic alignment and cross-modal semantic alignment, thereby enhancing the model's cross-modal consistency and instruction-following robustness.
Additionally, we present SemGen-Bench, a new benchmark specifically designed to evaluate multimodal semantic consistency under complex textual instructions.
Extensive experiments across multiple tasks and benchmarks demonstrate that UniAlignment outperforms existing baselines, underscoring the significant potential of diffusion models in unified multimodal generation. 
\end{abstract}


\section{Introduction}

The emergence of unified models for image generation and understanding~\cite{team2024chameleon, xie2024show, chen2025janus, pan2025transfer} reflects the growing demand for general-purpose generative frameworks in multimodal learning. 
Recent models, such as GPT-4o-Image~\cite{Gpt4o} and BLIP-3o~\cite{chen2025blip3} have demonstrated impressive capabilities in both visual comprehension and generation, suggesting the feasibility of integrating diverse vision-language tasks within a single architecture. 


Despite these advances, achieving a fully unified multimodal framework remains challenging. 
AR-based unified models~\cite{wu2025janus, xie2024show, zhou2024transfusion}, built upon transformer architectures adapted from large language models~\cite{bai2023qwen, touvron2023llama}, exhibit strong compositional reasoning and instruction-following capabilities. 
However, these models typically depend on discrete token-based decoding, which inherently constrained their ability to capture fine-grained visual fidelity and continuous pixel-level manipulation necessary for high-quality image synthesis and editing. 
\begin{figure}[t]
    \centering
    \includegraphics[width=1.0\linewidth]{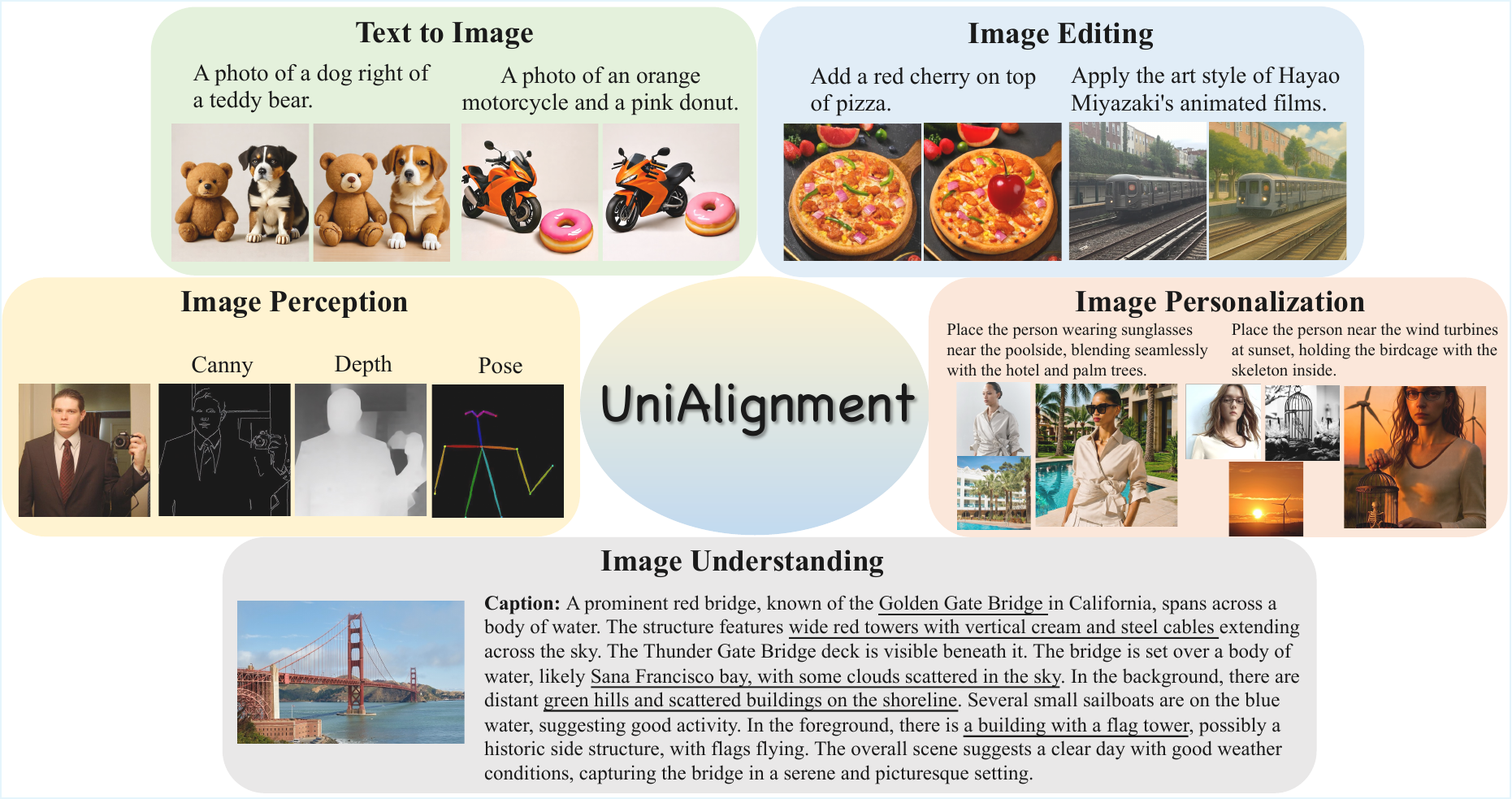}
    \vspace{-10pt}
    \caption{\textbf{Showcase of UniAlignment's capabilities.} Our approach enables a single lightweight DiT to handle diverse multimodal tasks, including image understanding, generation, editing, perception, and personalization, achieving versatile capabilities within a unified framework.}
    \vspace{-12pt}
    \label{fig:1}
\end{figure}
In contrast, Diffusion-based models~\cite{rombach2022high, peebles2023scalable, esser2024scaling, Flux} excel in generating high-fidelity images through iterative denoising, but often lack robust semantic perception.

Several methods employing hybrid architectures, such as MetaQueries~\cite{pan2025transfer},  Step1X-Edit~\cite{liu2025step1x}, UniWorld-V1~\cite{lin2025uniworld}, and OmniGen2~\cite{wu2025omnigen2}, which enhance diffusion with VLM guidance but remain structurally fragmented, involving modular designs and specialized connectors, thus diminishing their generalizability and scalability.
These approaches fail to construct a unified model architecture, resulting in redundant parameters that degrade both training and inference efficiency.
Recent efforts~\cite{li2025dual, swerdlow2025unified, yang2025mmada} attempt to construct a unified model purely based on diffusion models.
However, DualDiffusion~\cite{li2025dual} despite employing a unified diffusion transformer, suffers from task interference and weak semantic grounding. 
UniDisc~\cite{swerdlow2025unified} introduces a purely discrete diffusion approach, which enhances token-level alignment but limits pixel-level precision essential for fine-grained editing.
MMaDA~\cite{yang2025mmada} integrates reinforcement learning to align reasoning process, but depends on separate encoders, which increases complexity architectural and undermines end-to-end optimization. 


To address the aforementioned challenges, we propose UniAlignment, a unified multimodal generative framework based
entirely on a single shared diffusion transformer architecture, designed to harmonize image understanding, generation, editing, and perception tasks. 
The model adopts a dual-stream diffusion design, where a shared DiT backbone jointly parameterizes a continuous diffusion branch for image synthesis and a discrete diffusion branch for text-based understanding. This design enables a symmetric and scalable modeling of both modalities within the same generative space.
By avoiding reliance on external vision-language models (VLMs) and modular components during inference, UniAlignment maintains high computational efficiency and structural simplicity. 
To further enhance multimodal coherence, UniAlignment introduces two lightweight yet highly effective semantic alignment strategies. 
The first is cross-modal semantic alignment between text-to-image and image-to-text branches, corporating contrastive learning into dual stream training. 
The second is intrinsic-modal semantic alignment via intermediate feature matching, enriching the latent semantic during denoising process.
As shown in Fig.~\ref{fig:2}, these mechanisms significantly improve semantic consistency of instruction following and robustness in vision-language semantic grounding. 

To enable collaborative optimization across multiple tasks, we adopt a progressive multi-stage training strategy on large-scale datasets to achieve balanced performance.
Furthermore, given the limitations of existing benchmarks, which typically adopt simple semantics and rigid instruction formats, we construct a new benchmark, SemGen-Bench, specifically designed to evaluate models’ semantic fidelity and multimodal alignment under complex, compositional instructions.
Extensive experiments demonstrate that UniAlignment achieves remarkable performance across diverse multimodal tasks, outperforming existing unified models in both general scenarios and challenging generation and editing tasks. 
Our contributions can be summarized as follows: 

\begin{itemize}
\item[$\bullet$] 
We propose UniAlignment, a unified multimodal generative model based on a single Diffusion Transformer, demonstrating outstanding performance while maintaining lightweight design and computational efficiency.
\item[$\bullet$] 
We introduce two complementary semantic alignment mechanisms that significantly enhances image-text semantic consistency and instruction-following robustness.
\item[$\bullet$] 
We construct a rigorous new benchmark SemGen-Bench for evaluating multimodal semantic alignment under complex, compositional instructions, establishing a high-standard baseline for future research.
\end{itemize}


\section{Related Works}

\subsection{Semantic Representation Alignment}

Semantic representation alignment plays a pivotal role in vision-language representation learning, significantly influencing downstream tasks such as visual understanding and generation. 
Early works~\cite{radford2021learning, sun2023eva, zhai2023sigmoid} leverage large-scale image-text pairs and contrastive learning to construct multimodal pretraining frameworks that align visual and textual modalities. 
Other approaches~\cite{li2022blip, li2023blip} introduce Q-former~\cite{zhang2024vision} to bridge vision encoders and language models, achieving parameter-efficient multimodal adaptation. 
Building on these foundations, several methods focus on optimizing generative models through semantic representation alignment. 
~\cite{hudson2024soda, wangdiffusion, ma2025genhancer} utilize diffusion feedback to refine CLIP representations, aligning pretrained VLMs with diffusion-based generation through self-supervised reconstruction. 
Others aim to enhance perceptual quality by aligning internal representations between diffusion and vision models. 
Early works~\cite{xiang2023denoising, wei2023diffusion, tian2024addp} explore unified visual representation learning via self-supervised diffusion training. 
VAVAE~\cite{yao2025reconstruction} aligns high-dimensional visual tokens with backbone encoder features to push the limits of reconstruction quality, while REPA~\cite{yurepresentation} introduces intermediate-layer alignment between DiTs and pretrained vision encoders~\cite{caron2021emerging}, improving convergence and generation fidelity. 
Extending this idea, SoftREPA~\cite{lee2025aligning} employs contrastive learning with learnable soft tokens to enhance text-image consistency. 
In this work, we propose two semantic alignment strategies, jointly optimizing semantic consistency and visual-language grounding within a unified generative framework.

\subsection{Unified Multi-Modal Generation Models}

Recent breakthroughs in vision-language models~\cite{Gemini, Gpt4o, chen2025blip3} have driven a new wave of innovation in multimodal large models, accelerating the pursuit of unified frameworks for general-purpose generation. 
Early explorations~\cite{wang2024emu3, team2024chameleon, xie2024show, wu2025janus} focus on augmenting LLMs with visual encoders, using autoregressive decoding over pixel-level tokens to enable vision synthesis. 
Follow-up works~\cite{wu2024vila, qu2025tokenflow, xie2024muse} investigate the impact of discrete or continuous visual tokenization on unified vision-language modeling, still grounded within autoregressive transformer architectures. 
In contrast, methods~\cite{li2025dual, swerdlow2025unified, yang2025mmada} attempt to unify image generation and understanding via diffusion models, proposing dual-stream diffusion frameworks that jointly model denoising processes in discrete or continuous latent spaces.

However, achieving general-purpose multimodal generative models requires going beyond text-to-image synthesis. 
A line of recent work~\cite{brooks2023instructpix2pix, zhang2023magicbrush, yu2025anyedit, shi2024seededit, wei2024omniedit, wang2025mind, liu2025step1x} extends the capabilities of diffusion models to support instruction-based image editing. 
More recent efforts~\cite{mao2025ace++, xiao2025omnigen, zhang2025nexus, pan2025transfer, ai2025ming, huang2025illume+, lin2025uniworld} aim to build multitask, generalizable visual generation frameworks. 
Among them,~\cite{pan2025transfer, ai2025ming, huang2025illume+, lin2025uniworld} adopt a hybrid architecture, using connectors to bridge VLMs and diffusion backbones, thereby combining semantic understanding with pixel-level synthesis. 
Although representing a step towards integrating visual understanding, generation, and manipulation, these methods remain structurally fragmented, resulting in redundant computation and a lack of architectural unification.

\begin{figure}[t]
    \centering
    \vspace{-6pt}
    \includegraphics[width=1.0\linewidth]{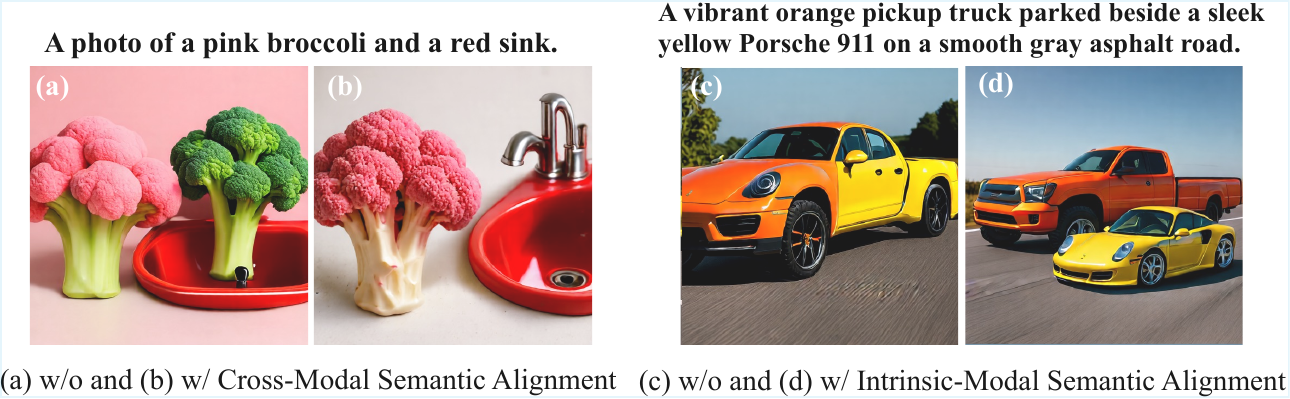}
    \vspace{-12pt}
    \caption{\textbf{Analysis of the proposed semantic representation alignment.} (a–d) present the generated images with and without the two semantic alignment mechanisms.}
    \vspace{-12pt}
    \label{fig:2}
\end{figure}

\section{Method}
In this section, we present UniAlignment, a unified framework for multimodal generation tasks.
We begin by introducing the overall architecture, which jointly addresses both image generation and understanding within a single diffusion-based backbone. 
To meet the semantic alignment demands posed by various vision-language generation scenarios, we propose two lightweight yet effective semantic alignment strategies: cross-modal semantic alignment and intrinsic-modal semantic alignment, implicitly enhancing the semantic grounding capability of diffusion models without additional parameters or learnable queries. 
To further support balanced learning across diverse generation tasks, a multi-stage training strategy is introduced to improve multi-task performance by progressively optimization.

\subsection{Unified Generation \& Understanding Framework}

Autoregressive models have demonstrated limited efficacy in visual generation tasks, struggling with fine-grained spatial coherence and semantic fidelity. 
In contrast, diffusion models have shown remarkable potential for general-purpose generation. 
Motivated by this, we leverage a single diffusion transformer (DiT) to unify both visual generation and visual understanding, without relying on VLMs or external visual encoders during inference.

\begin{figure}[t]
    \centering
    \vspace{-6pt}
    \includegraphics[width=1.0\linewidth]{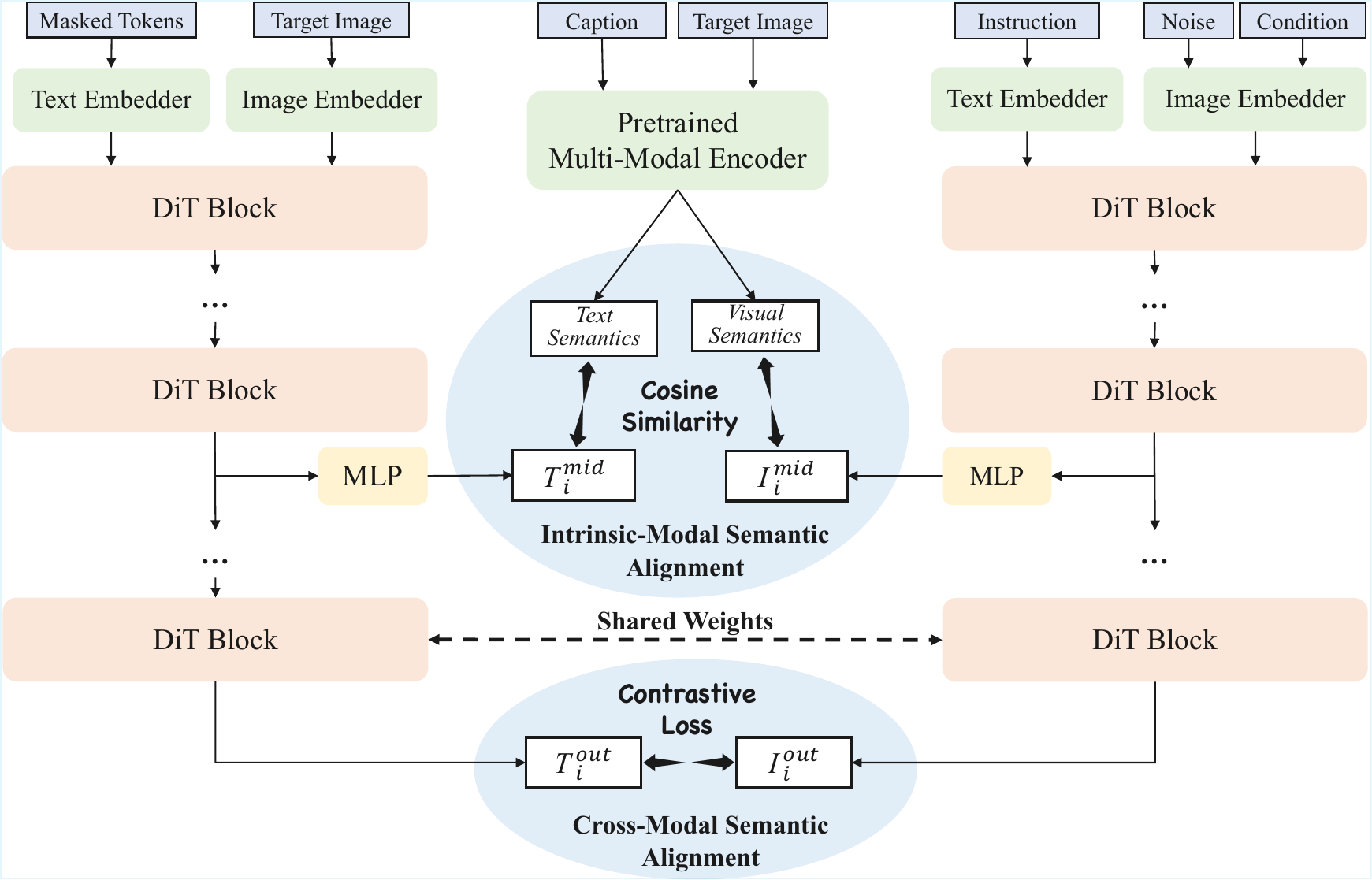}
    \vspace{-10pt}
    \caption{\textbf{The training pipeline of UniAlignment.} Text diffusion (left branch) and image diffusion (right branch) are both unified within a single Transformer with shared weights. The central part illustrates the proposed cross-modal semantic alignment and intrinsic-modal semantic alignment. The image condition is set absent during T2I generation. Only parameters within the DiT blocks and the MLP heads are optimized throughout the entire training.}
    \vspace{-12pt}
    \label{fig:3}
\end{figure}

As illustrated in Fig.~\ref{fig:3}, UniAlignment adopts a dual-stream diffusion architecture, parameterizing both continuous and discrete diffusion processes within a shared DiT backbone. 
Inspired by DualDiffusion~\cite{li2025dual}, this architecture simultaneously models image distribution $p(\mathbf{x}_\mathrm{img})$ and text distribution $p(\mathbf{x}_\mathrm{txt})$. 
The continuous diffusion branch targets image generation following the flow matching principle~\cite{lipmanflow}, with its training objective defined as follows:
\begin{equation}
\mathcal{L}_{\text{img}} =\mathbb{E}_{t, q_{\mathrm{img}}}\left\|\mathbf{v}_\theta\left(\mathbf{x}^t_{\mathrm{img}}, t, \mathbf{x}_{\mathrm{txt}}\right)-\left(\boldsymbol{\epsilon}-\mathbf{x}_{\mathrm{img}}\right)\right\|_2^2,
\end{equation}
where $\mathbf{x}^t_{\mathrm{img}}$ denotes the noisy image at timestep $t$, $\mathbf{v}_\theta$ is the predicted velocity field, and $\boldsymbol{\epsilon}$ represents the noise. 

The discrete diffusion branch, masked token prediction is adopted to model the reverse process based on the masked state of discrete tokens. 
Specifically, the model learns to iteratively denoise masked tokens, thereby approximating the underlying data distribution in the discrete space:
\begin{equation}
\mathcal{L}_{\text{txt}} =\mathbb{E}_{q_{\mathrm{txt}}}\left[-\frac{1}{K} \sum_{i=1}^K \log \left[\mathbf{x}_\theta\left(\mathbf{x}^{t_i}_{\mathrm{txt}}, \mathbf{x}_{\mathrm{img}}\right) \cdot \mathbf{x}\right] / t_i\right],
\end{equation}
where timestep $t_i$ is determined based on the text sequence length. 
The understanding and generation branches share the same parameters throughout training, allowing both modalities to be jointly optimized within a unified architecture.

To support general-purpose multimodal generation, the continuous diffusion branch extends beyond text-to-image synthesis to include a broad range of conditional image generation tasks, including image manipulation (e.g., instruction-based editing, subject customization, stylization) and image perception (e.g., depth estimation, edge detection, and human pose prediction). 
The source images are first encoded via a pretrained VAE to obtain conditional tokens, which are then concatenated with the noisy latents and fed into the DiT blocks. 
During training, the continuous diffusion branch receives both the conditional image and the textual instruction as input, while the discrete diffusion branch is conditioned on the target image and its corresponding caption. 
This design enables the incorporation of visual conditions without architectural modifications, additional encoders, or query tokens. 
The unified DiT framework of UniAlignment allows seamless extension to a variety of multimodal generation tasks, maintaining lightweight design and high efficiency, avoiding the semantic gap between autoregressive and diffusion-based paradigms.

\subsection{Semantic Representation Alignment}
\subsubsection{Cross-modal Semantic Alignment.}

Although jointly training bi-directional diffusion processes within a single transformer allows for synchronized learning of generation and understanding, the inherent differences in their optimization objectives can lead to conflicting gradients and suboptimal convergence. 
As illustrated in Fig.~\ref{fig:2}(a-b), these conflicts often degrade instruction-following capabilities, reducing the model's text-image consistency. 
This highlights the challenge of achieving balanced optimization in unified multimodal frameworks.

To alleviate the conflict arising from bi-directional diffusion training, we propose a solution that adjusts the direction of joint optimization. 
Our key insight is to leverage the benefits of joint vision-language training to improve text-image consistency through representation alignment within the transformer. 
As contrastive learning has proven effective in aligning multimodal representations in prior works~\cite{radford2021learning, li2022blip, jia2021scaling}, we introduce a contrastive objective over the output embeddings from the dual diffusion branches. This objective encourages semantic similarity between matched image-text pairs while suppressing that of unmatched pairs. 
Specifically, given a batch of $N$ image-text pairs,we optimize the contrastive loss over the corresponding embeddings to strengthen cross-modal semantic alignment as follows:
\begin{equation}
\mathcal{L}_{\mathrm{Cross}}=-\frac{1}{N} \sum_{i=1}^N \log \frac{\exp \left(\operatorname{sim}\left(\mathbf{I}^{o}_i, \mathbf{T}^{o}_i\right) / \tau\right)}{\sum_{j=1}^N \exp \left(\operatorname{sim}\left(\mathbf{I}_i, \mathbf{T}_j\right) / \tau\right)},
\end{equation}
where $\mathbf{I}^{o}_i$ and $\mathbf{T}^{o}_i$ denote the output embeddings from the image and text diffusion branches, respectively. $\tau$ is the temperature parameter. 

\subsubsection{Intrinsic-modal Semantic Alignment.}

Unified multimodal generation poses high requirements on a model’s representational capacity, particularly in tasks such as image editing and visual perception. 
These tasks require the model to not only comprehend textual semantics but also accurately perceive high-level visual attributes from images. 
However, the T5 text encoder and the VAE used in DiT models are pretrained independently on unimodal data, lacking the ability to capture the inherent semantic dependencies between modalities.
As a result, the model struggles with aligning text and image semantics effectively.
As shown in Fig.~\ref{fig:2}(c-d), current DiT-based model exhibit limited semantic understanding, often failing to detect critical attributes embedded in the textual instruction and image content.

To address the this limitation, we propose an intrinsic-modal semantic alignment strategy that enhances the model’s visual-linguistic perception without introducing additional inference-time parameters. 
Inspired by REPA~\cite{yurepresentation}, we regularize  intermediate hidden states of the DiT blocks using semantic embeddings extracted from a pretrained vision-language encoder. 
Specifically, for each image-text pair, the target image and caption are passed through a pretrained vision-language encoder to obtain semantic embeddings $\mathbf{h}_I$ and $\mathbf{h}_T$. 
These embeddings are then aligned with the hidden states $\mathbf{I}^{m}_i$ and $\mathbf{I}^{m}_t$ from specific layers of the model's image and text diffusion branches, respectively. 
Two Multilayer perceptrons (MLPs) are introduced as projection heads $M$ to mitigate the dimension discrepancy. 
The intrinsic-modal semantic alignment loss is defined as:
\begin{equation}
\scalebox{0.95}{$
\begin{aligned}
& \mathcal{L}_{\mathrm{Intrinsic}} \\
& =-\frac{1}{N}\sum_{n=1}^N \left[\operatorname{sim}\left(\mathbf{h}_I, M\left(\mathbf{I}^{m}_i\right)\right)+\operatorname{sim}\left(\mathbf{h}_T, M\left(\mathbf{T}^{m}_i\right)\right)\right],
\end{aligned}
$}
\end{equation}
where $\operatorname{sim}$ denotes the cosine similarity. Notably, both the pretrained vision-language encoder and the MLPs are only used during training and do not introduce any additional parameters during inference. 

\begin{figure}[t]
    \centering
    \includegraphics[width=1.0\linewidth]{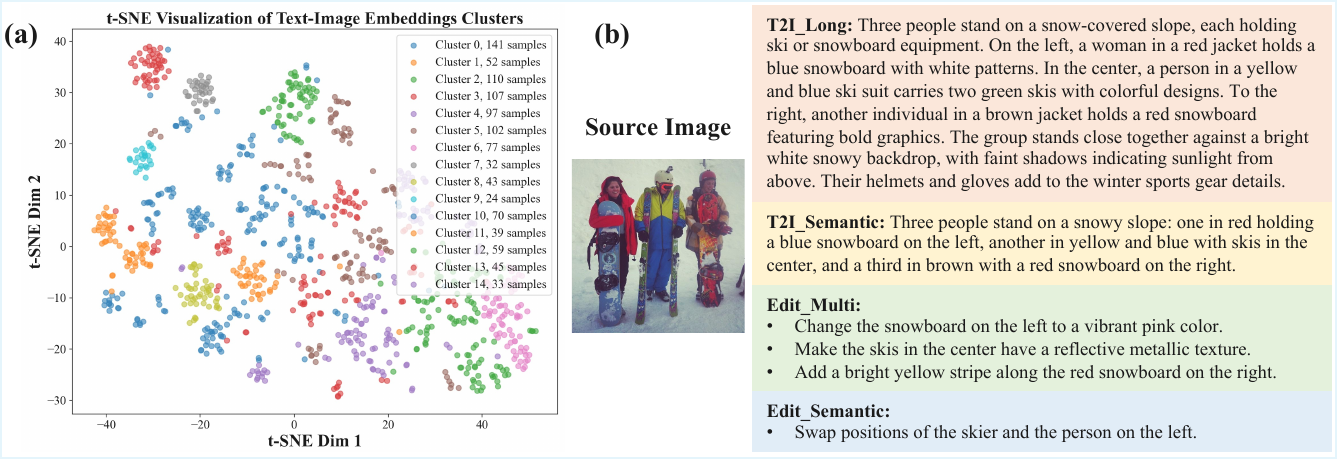}
    \vspace{-12pt}
    \caption{\textbf{The proposed SemGen-Bench.} (a) Visualization of clustered embeddings of image data. (b) Samples from the proposed SemGen-Bench.}
    \vspace{-10pt}
    \label{fig:4}
\end{figure}

\begin{table*}
    \centering
    \setlength{\tabcolsep}{4.5pt}
    \vspace{-8pt}
    \caption{\textbf{Comparison across Generation, Editing and Understanding tasks.} *: The first term and the second term represent the number of parameters for text generation and image generation, respectively. † refers to the methods using LLM rewriter.} 
    \vspace{-6pt}
    \label{table:1}
{\tiny
\begin{tabular}{llcccccccccccc}
\toprule
\multirow{2}{*}{Type} & \multirow{2}{*}{Method} &  \multirow{2}{*}{Params}  & \multicolumn{2}{c}{T2I Generation}    & \multicolumn{2}{c}{Image Editing} & \multicolumn{3}{c}{Understanding} & \multicolumn{4}{c}{SemGen-Bench} \\   
\cmidrule(r){4-5}   \cmidrule(r){6-7}  \cmidrule(r){8-10} \cmidrule(r){11-14}
 & & & GenEval & DPG-Bench & ImgEdit-Bench & GEdit-Bench-EN & MMB & MMMU & MM-Vet & T2I\_{Long} & T2I\_{Sem} & Edit\_{Multi} & Edit\_{Sem} \\
\cmidrule(r){1-5}  \cmidrule(r){6-7}   \cmidrule(r){8-10} \cmidrule(r){11-14} 
\multirow{3}{*}{Gen. Only} &
SDXL         & - & 0.55 & 74.7 & - & - & - & - & - & - & - & - & -  \\ 
& SD3-medium   & - & 0.62 & 84.1 & - & - & - & - & - & - & - & - & - \\ 
& FLUX.1-dev   & - & 0.66 & 84.0 & - & - & - & - & - & - & - & - & - \\
& DualDiffusion  & - & 0.65 & 81.3 & - & - & - & - & - & 7.71 & 7.29 & - & - \\
\cmidrule{1-14}
\multirow{5}{*}{Edit. Only} &
Instruct-P2P & - & - & - & 1.88 & 3.68 & - & - & - & - & - & - & - \\
& MagicBrush   & - & - & - & 1.90 & 1.86 & - & - & - & - & - & - & - \\
& AnyEdit      & - & - & - & 2.45 & 3.21 & - & - & - & - & - & - & -\\
& Step1X-Edit  & - & - & - & 3.06 & \textbf{6.70} & - & - & - & - & - & 6.24 & \textbf{6.10} \\
& IC-Edit      & - & - & - & 3.05 & 4.84 & - & - & - & - & - & - & -\\
\cmidrule{1-14}
\multirow{2}{*}{Und. Only} &
LLaVA-1.5   & - & - & - & - & - & 36.4 & \textbf{67.8} & 36.3 & - & - & - & -  \\
& LLaVA-NeXT & - & - & - & - & - & 79.3 & 51.1 & 57.4 & - & - & - & -  \\
\cmidrule{1-14}
\multirow{7}{*}{Unified} &
Show-o       & - & 0.68 & 67.27 & - & - & - & 27.4 & - & - & - & - & -  \\
& Janus-Pro    & - & 0.80 & 84.19 & - & - & 75.5 & 36.3 & 39.8 & - & - & - & - \\
& Emu3         & - & 0.54 & 80.60 & - & - & 58.5 & 31.6 & 37.2 & - & - & - & -  \\
& BAGEL        & 7B+7B$^{*}$ & \textbf{0.82} & \underline{85.07} & 3.20 & 6.52 & \textbf{85.0} & 55.3 & \textbf{67.2} & - & - & - & - \\
& Uniworld-V1  & 7B+12B$^{*}$ & 0.80 & 81.38 & \underline{3.26} & 4.85 & \underline{83.5} & 58.6 & \underline{67.1} & 6.42 & 5.58 & 4.76 & 5.77 \\
& OmniGen2     & 3B+4B$^{*}$ & 0.77 & 83.57 & \textbf{3.44} & 6.42 & 79.1 & 53.1 & 61.8 & \underline{7.95} & \underline{7.90} & \underline{6.40} & 5.61 \\
\cmidrule{2-14}
& \textbf{UniAlignment} & \textbf{2B} & \underline{0.81} & \textbf{85.64} & 3.25 & \underline{6.57} & 80.6 & \underline{61.3} & 63.0 & \textbf{8.07} & \textbf{7.97} & \textbf{6.85} & \underline{5.98} \\
 \bottomrule
\end{tabular}
}
\vspace{-4pt}
\end{table*}

Based on aforementioned designs, the overall optimization objective is the weighted sum of above loss functions:
\begin{equation}
    \mathcal{L} = \mathcal{L}_{img} + \mathcal{\lambda}_1\mathcal{L}_{txt} + \mathcal{\lambda}_2\mathcal{L}_{\mathrm{Cross}} + \mathcal{\lambda}_3\mathcal{L}_{\mathrm{Intrinsic}}
\end{equation}
The integration of semantic alignment facilitates fine-grained semantic understanding of the lightweight DiT framework and improves the model’s ability to capture cross-modal dependencies across diverse generation tasks.

\subsection{Multi-Stage Training}

To effectively train a unified generative model across diverse tasks and modalities, we design a multi-stage training strategy that promotes task balance and gradual performance improvement. 
In the first stage, we conduct image-text pretraining using 2 million image-text pairs from Text-to-Image 2M~\cite{T2I2M} and an additional 10 million curated internal samples, covering both captioning and generation tasks. This stage focuses on building foundational alignment for both captioning and generation tasks.
The second stage is multi-task joint pretraining.
We incorporate heterogeneous datasets, including image editing and visual perception examples collected from previous works~\cite{yu2025anyedit, zhang2023magicbrush, shi2024seededit, wei2024omniedit, yu2024promptfix, lin2025uniworld}, as well as GPT-4o-generated samples. 
To adapt these datasets for text-image joint training, we use vision-language models to generate captions for target images lacking textual descriptions.
To prevent task-specific overfitting and catastrophic forgetting, T2I samples are interleaved with conditional task data during this stage. 
A hybrid batch sampling strategy is adopted, where multiple data types are alternated within training iterations, enabling balanced optimization across tasks. 
The final stage is supervised finetuning using high-quality datasets including BLIP-3o~\cite{chen2025blip3}, ShareGPT-4o~\cite{chen2025sharegpt}, and LLaVA-OneVision~\cite{li2024llava}.
The progressive multi-stage training encourages the model to incrementally acquire multimodal competencies while mitigating  optimization imbalance across diverse tasks.

\section{Experiments}
\subsection{Implementation Details}

\begin{figure}[t]
    \centering
    \includegraphics[width=1.0\linewidth]{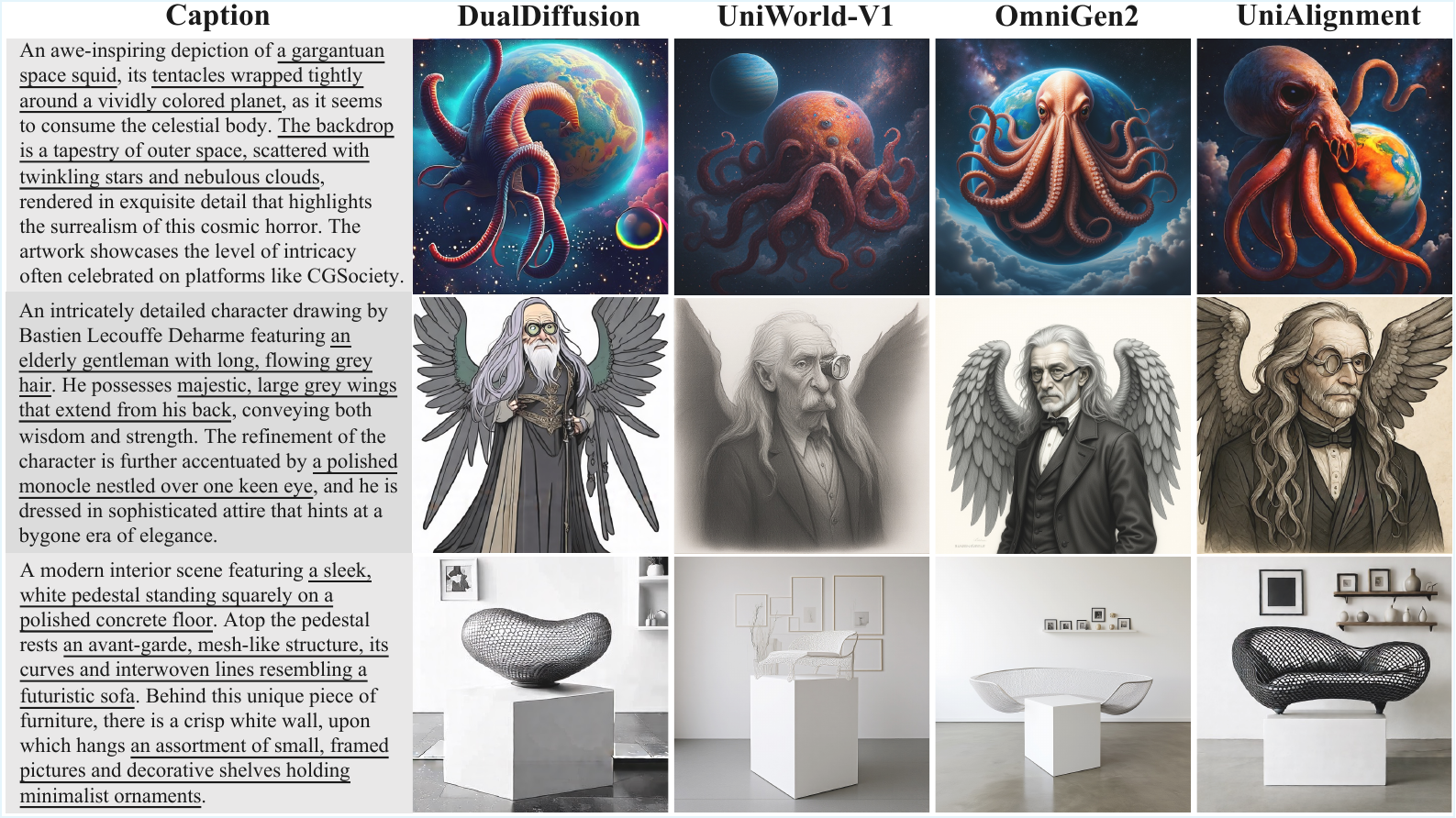}
    \vspace{-12pt}
    \caption{\textbf{Comparison of text to image generation with opensource approaches.} The underlined parts emphasize the details in the text instructions.}
    \vspace{-12pt}
    \label{fig:5}
\end{figure}

\begin{figure}[t]
    \centering
    \includegraphics[width=1.0\linewidth]{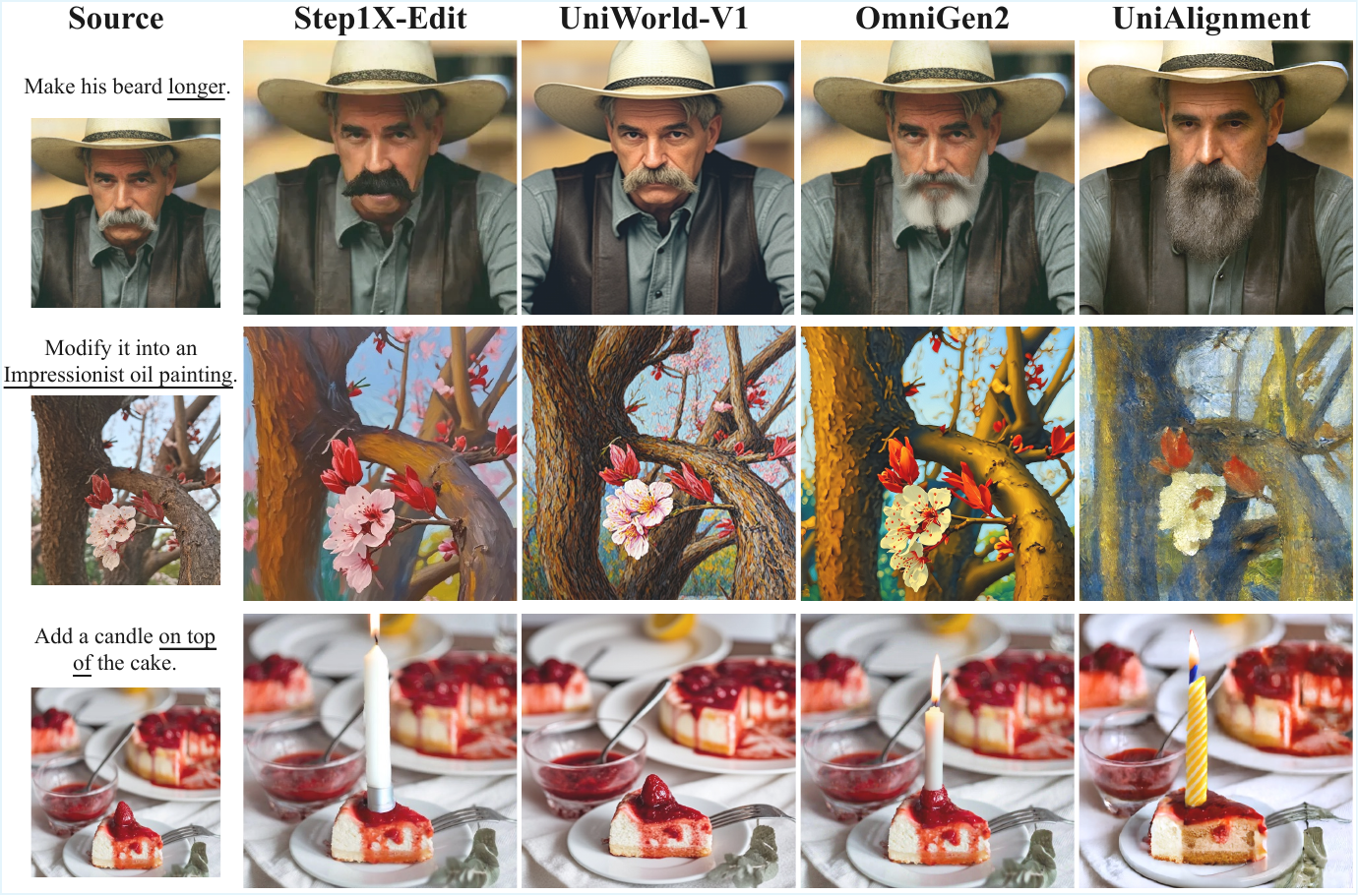}
    \vspace{-12pt}
    \caption{\textbf{Comparison of instruction-guided image editing.}}
    \vspace{-12pt}
    \label{fig:6}
\end{figure}

\begin{table*}
    \centering
    \vspace{-6pt}
    \caption{\textbf{Evaluation of text-to-image generation ability on GenEval benchmark and DPG-Bench.}} 
    \vspace{-6pt}
    \label{table:2}
{\tiny
\begin{tabular}{lccccccccccccc}
\toprule
\multirow{2}{*}{Method} &  \multicolumn{7}{c}{GenEval}  & \multicolumn{6}{c}{DPG-Bench}  \\   
\cmidrule(r){2-8}   \cmidrule(r){9-14} 
 & Single object & Two object  & Counting & Colors & Position & Color attribution & Overall & Global & Entity & Attribute & Relation & Other & Overall \\  
  \cmidrule{1-14} 
DualDiffusion & 0.97 & 0.80 & 0.54 & 0.76 & 0.32 & 0.50 & 0.65 & 87.33 & 89.48 & 86.72 & 89.95 & 86.32 & 81.32 \\ 
SDXL         & 0.98 & 0.74 & 0.39 & 0.85 & 0.15 & 0.23 & 0.55 & 83.27 & 82.43 & 80.91 & 86.76 & 80.41 & 74.65 \\ 
SD3-medium   & 0.99 & 0.94 & 0.72 & 0.89 & 0.33 & 0.60 & 0.74 & 87.90 & 91.01 & 88.83 & 80.70 & 88.68 & 84.08 \\ 
FLUX.1-dev   & 0.99 & 0.81 & 0.79 & 0.74 & 0.20 & 0.47 & 0.67 & 82.10 & 89.50 & 88.70 & 91.10 & 89.40 & 84.00 \\
LUMINA-Next  & 0.92 & 0.46 & 0.48 & 0.70 & 0.09 & 0.13 & 0.46 & 82.82 & 88.65 & 86.44 & 80.53 & 81.82 & 74.63 \\
OmniGen      & 0.98 & 0.84 & 0.66 & 0.74 & 0.40 & 0.43 & 0.68 & 87.90 & 88.97 & 88.47 & 87.95 & 83.56 & 81.16 \\
\cmidrule{1-14}
Show-o       & 0.98 & 0.80 & 0.66 & 0.84 & 0.31 & 0.50 & 0.68 & 79.33 & 75.44 & 78.02 & 84.45 & 60.80 & 67.27 \\
Janus        & 0.97 & 0.68 & 0.30 & 0.84 & 0.46 & 0.42 & 0.61 & 82.33 & 87.38 & 87.70 & 85.46 & 86.41 & 79.68 \\
Janus-Pro    & 0.99 & 0.89 & 0.59 & 0.90 & 0.79 & 0.66 & 0.80 & 86.90 & 88.90 & 89.40 & 89.32 & 89.48 & 84.19 \\
Emu3         & 0.99 & 0.81 & 0.42 & 0.80 & 0.49 & 0.45 & 0.66 & 85.21 & 86.68 & 86.84 & 90.22 & 83.15 & 80.60 \\
TokenFlow-XL & 0.95 & 0.60 & 0.41 & 0.81 & 0.16 & 0.24 & 0.55 & 78.72 & 79.22 & 81.29 & 85.22 & 71.20 & 73.38 \\
BAGEL        & 0.99 & 0.94 & 0.81 & 0.88 & 0.64 & 0.63 & \textbf{0.82} & 88.94 & 90.37 & 91.29 & 90.82 & 88.67 & \underline{85.07} \\
Uniworld-V1  & 0.99 & 0.93 & 0.79 & 0.89 & 0.49 & 0.70 & 0.80 & 83.64 & 88.39 & 88.44 & 89.27 & 87.22 & 81.38 \\
OmniGen2     & 1 & 0.94 & 0.64 & 0.88 & 0.45 & 0.70 & 0.77 & 88.81 & 88.83 & 90.18 & 89.37 & 90.27 & 83.57 \\
\cmidrule{1-14}
\textbf{UniAlignment} & 0.99 & 0.95 & 0.76 & 0.88 & 0.56 & 0.78 & \underline{0.81} & 91.84 & 90.16 & 90.44 & 91.58 & 89.55 & \textbf{85.64} \\
 \bottomrule
\end{tabular}
}
\vspace{-1pt}
\end{table*}


We build UniAlignment upon the open-sourced SD 3.0 backbone~\cite{esser2024scaling}. A frozen T5 text encoder is employed, and image tokenization is performed using a pretrained VAE.
The core architecture, a multimodal DiT, contains only 2 million trainable parameters.
The vision encoder of Qwen2.5-VL-7B is utilized for intrinsic-modal semantic alignment instead of CLIP or SigLIP due to their limited input length for text tokens. 
The DiT embeddings are sourced from a transformer block at depth 8, which we empirically found to yield optimal performance.

Training images are resized to 512 × 512, and text sequences are truncated to 256 tokens. 
The model is trained for 80K steps in the first two stages and 30K steps in the finetuning stage, totaling 25M training samples. 
During training, we adopt a constant learning rate of 3e-5 and weight decay schedule of 1e-2. 
Gradient accumulation is employed throughout all training stages, allowing for a total batch size of 256. 
The overall loss function parameters are set as $\lambda_1=0.2$, $\lambda_2=0.05$ and $\lambda_3=0.1$. 

\begin{table*}
\setlength{\tabcolsep}{9pt}
    \centering
    \caption{\textbf{Quantitative comparison on GEdit-Bench-EN and ImgEdit-Bench.} For GEdit-Bench, SC (Semantic Consistency) evaluates instruction following, and PQ (Perceptual Quality) assesses image naturalness and artifacts. Higher scores are better.} 
    \vspace{-4pt}
    \label{table:3}
{\tiny
\begin{tabular}{lccccccccccccc}
\toprule
\multirow{2}{*}{Method} &  \multicolumn{3}{c}{Gedit-Bench-EN}  & \multicolumn{10}{c}{ImgEdit-Bench}  \\   
\cmidrule(r){2-4}   \cmidrule(r){5-14} 
 & SC & PQ  & O & Add & Adjust & Extract & Replace & Remove & Background & Style & Hybrid & Action & Overall \\  
\cmidrule{1-14}
Instruct-P2P & 3.58 & 5.49 & 3.68 & 2.45 & 1.83 & 1.44 & 2.01 & 1.50 & 1.44 & 3.55 & 1.2 & 1.46 & 1.88 \\ 
MagicBrush   & 4.68 & 5.66 & 4.52 & 2.84 & 1.58 & 1.51 & 1.97 & 1.58 & 1.75 & 2.38 & 1.62 & 1.22 & 1.90  \\
AnyEdit      & 3.18 & 5.82 & 3.21 & 3.18 & 2.95 & 1.88 & 2.47 & 2.23 & 2.24 & 2.85 & 1.56 & 2.65 & 2.45 \\
OmniGen      & 5.96 & 5.89 & 5.06 & 3.47 & 3.04 & 1.71 & 2.94 & 2.43 & 3.21 & 4.19 & 2.24 & 3.38 & 2.96 \\
Step1X-Edit  & 7.09 & 6.76 & \textbf{6.70} & 3.88 & 3.14 & 1.76 & 3.40 & 2.41 & 3.16 & 4.63 & 2.64 & 2.52 & 3.06 \\
ICEdit       & 5.11 & 6.85 & 4.84 & 3.58 & 3.39 & 1.73 & 3.15 & 2.93 & 3.08 & 3.84 & 2.04 & 3.68 & 3.05 \\
BAGEL        & 7.36 & 6.83 & 6.52 & 3.56 & 3.31 & 1.70 & 3.30 & 2.62 & 3.24 & 4.49 & 2.38 & 4.17 & 3.20 \\
Uniworld-V1  & 4.93 & 7.43 & 4.85 & 3.82 & 3.64 & 2.27 & 3.47 & 3.24 & 2.99 & 4.21 & 2.96 & 2.74 & \underline{3.26} \\
OmniGen2     & 7.16 & 6.77 & 6.41 & 3.57 & 3.06 & 1.77 & 3.74 & 3.20 & 3.57 & 4.81 & 2.52 & 4.68 & \textbf{3.44} \\
\cmidrule{1-14}
\textbf{UniAlignment} & 7.25 & 6.81 & \underline{6.57} & 3.66 & 3.21 & 2.07 & 3.45 & 2.51 & 3.28 & 4.63 & 2.50 & 3.95 & 3.25\\
 \bottomrule
\end{tabular}
}
\vspace{-8pt}
\end{table*}

\subsection{Evaluation Settings}
To comprehensively evaluate UniAlignment, we access its performance across multiple standard and newly proposed benchmarks. 
Specifically, text-to-image generation is assessed using GenEval~\cite{ghosh2023geneval} and DPG-Bench~\cite{hu2024ella}, while image editing capabilities are evaluated on GEdit-Bench-EN~\cite{liu2025step1x} and ImgEdit-Bench~\cite{ye2025imgedit}. To further investigate multimodal semantic fidelity under complex compositional instructions, we introduce SemGen-Bench. Qualitative assessments on tasks such as image perception and personalization further indicate the versatility of our approach.

\subsection{Main Results}
As shown in Table~\ref{table:1}, we conduct a comprehensive comparison of our proposed UniAlignment with state-of-the-art models across text-to-image generation, image editing, visual understanding, and the proposed SemGen-Bench. 
Despite its compact scale (2B parameters), UniAlignment achieves comparable overall performance across all tasks. These results demonstrate the effectiveness of our unified DiT architecture and semantic alignment strategies in enabling a lightweight, general-purpose multimodal framework.
Below, we detail the evaluation results for each task.


\subsection{Detailed Task-specific Evaluation}
\subsubsection{New Evaluation Benchmark: SemGen-Bench}
A critical challenge in evaluating unified generative models lies in assessing their ability of modelling multimodal semantics, particularly under complex semantic instructions. 
Although taking visual attributes into account, existing visual generation benchmarks primarily focus on single image synthesis tasks, often under simplistic and narrowly defined settings. 
For instance, Geneval~\cite{ghosh2023geneval} evaluates compositional subject attributes within images, but it restricts instruction length and format to predefined templates, limiting its applicability to arbitrary textual inputs. 
GEdit-Bench~\cite{liu2025step1x} and ImgEdit-Bench~\cite{ye2025imgedit} target real-world image editing but confined to single-turn instructions with limited semantic depth. 
Prompts in these benchmarks are often direct and lack the complexity required to evaluate fine-grained semantic perception. 
To address these limitations, we introduce \textbf{SemGen-Bench (Semantic Generation Benchmark)}, a large-scale benchmark designed to measure model's visual generation performance in semantically challenging scenarios.
It spans both image generation and editing tasks under diverse and semantically rich instructions, providing a comprehensive assessment of multimodal generative models.

As illustrated in Fig.~\ref{fig:4}, SemGen-Bench comprises four task categories: (1)\textit{T2I\_ Long} (long-form text-to-image generation) includes instructions exceeding 200 tokens, testing the ability to process extended contextual information. (2)\textit{T2I\_ Sem} (semantically complex text-to-image generation) demands descriptions incorporating object count, color, spatial relationships, background attributes and logical reasoning. (3)\textit{Edit\_ Multi} (multi-turn image editing) involves three parallel editing instructions applied to a single image, evaluating the ability to execute coordinated transformations. (4)\textit{Edit\_ Sem} (semantically complex image editing) focuses on modifying object-level attributes while requiring semantic reasoning beyond surface-level changes.
Each category contains 150 examples covering a wide range of entities and scenes. 

To construct SemGen-Bench, we adopt a hybrid pipeline combining MLLMs with human annotation. 
Raw images from MSCOCO~\cite{lin2014microsoft} are filtered to remove samples with low quality and extreme aspect ratios. 
Next, a multimodal embedding model~\cite{zhang2024gme} is employed to extract vector representations of text-image pairs, followed by hierarchical clustering to obtain 15 semantically diverse categories. 
Image samples are selected from each category to ensure diversity and balance in the test set. 
Prompt instructions are first generated using Qwen2.5-VL-72B~\cite{bai2025qwen2}, then manually refined to ensure broad diversity in both semantic content and syntactic structure.

To ensure fair evaluation under complex semantic conditions, we adopt an MLLM-based protocol to assess model outputs across three dimensions: Instruction Consistency (IC), Perceptual Quality (PQ), and an Overall Score.
Following the VIEScore framework~\cite{ku2024viescore}, outputs are rated on a 0–10 scale using MLLMs. 
The SemGen-Bench will serve as a valuable benchmark for advancing research in unified multimodal generation.

\subsubsection{Text-to-Image Generation.}
As shown in Table~\ref{table:2}, UniAlignment achieves highly competitive performance with state-of-the-art methods across all evaluated subcategories.
Notably, UniAlignment achieves the highest overall score on DPG-Bench demonstrating superior capability in handling dense textual prompts. Additionally, UniAlignment secures the second-highest score on GenEval with a marginal gap of only 0.01, underscoring its robustness in semantically understanding visual attributes.
Qualitative comparisons with three unified generation methods~\cite{li2025dual, lin2025uniworld, wu2025omnigen2} are presented in Fig.~\ref{fig:5}. 
The visualization results exhibit stronger consistency with textual prompts of UniAlignment compared to existing methods, while also achieving superior visual fidelity.

\begin{table}
    \setlength{\tabcolsep}{2.5pt}
    \centering
    \vspace{-6pt}
    \caption{\textbf{Overall comparison on our proposed SemGen-Bench.} Instruction Consistency (IC), Perceptual Quality (PQ) and Overall Scores (O) are represented. ``Time" denotes the average generation time of each image.}
    \label{table:4}
    \vspace{-5pt}
{\tiny
    \begin{tabular}{@{}lccccccccccccccc@{}}
    \toprule
\multirow{2}{*}{Method} & \multirow{2}{*}{Time} & \multicolumn{3}{c}{T2I-Long} & \multicolumn{3}{c}{T2I-Sem} & \multicolumn{3}{c}{Edit-Multi} & \multicolumn{3}{c}{Edit-Sem}  \\   
\cmidrule(r){3-5} \cmidrule(r){6-8} \cmidrule(r){9-11} \cmidrule(r){12-14} 
 &  & IC & PQ & O & IC & PQ & O & IC & PQ & O & IC & PQ & O \\
    \midrule
    DualDiffusion & \textbf{2s} & 7.86 & 7.65 & 7.71 & 7.49 & 7.54 & 7.29 & - & - & - & - & - & - \\
    Step1X-Edit & 91s & - & - & - & - & - & - & 6.16 & 6.65 & 6.24 & \textbf{6.43} & \underline{6.84} & \textbf{6.10} \\
    UniWorld    & 21s  & 5.86 & \textbf{7.82} & 6.42 & 5.13 & 7.59 & 5.58 & 4.22 & \textbf{7.16} & 4.76 & 5.98 & \textbf{6.95} & 5.77 \\
    OmniGen2    &  37s  & \underline{8.29} & 7.72 & \underline{7.95} & \underline{8.29} & \textbf{7.77} & \underline{7.90} & \underline{6.42} & 6.72 & \underline{6.40} & 5.97 & 6.70 & 5.61 \\
    \textbf{UniAlignment} & \textbf{2s} & \textbf{8.53} & \underline{7.80} & \textbf{8.07} & \textbf{8.48} & \underline{7.69} & \textbf{7.97} & \textbf{6.94} & \underline{7.03} & \textbf{6.85} & \underline{6.23} & 6.80 & \underline{5.98}  \\
    \bottomrule
    \end{tabular}
}
\vspace{-10pt}
\end{table}

\subsubsection{Image Editing.}

As shown in Table~\ref{table:3}, UniAlignment ranks second only to the specialized editing method Step1X-Edit~\cite{liu2025step1x} on GEdit-Bench, outperforming UniWorld (1.72) and OmniGen2 (0.16). Notably, UniAlignment achieves the highest score in Semantic Consistency, highlighting its strong capability for joint visual-linguistic semantic grounding.
On ImgEdit-Bench, UniAlignment consistently ranks in the top three across nearly all metrics, demonstrating robust and consistent performance in instruction-based image editing tasks.
Qualitative visualizations with open-sourced approaches are provided in Fig.~\ref{fig:6}. Without relying on VLMs or additional visual encoders, UniAlignment demonstrates superior editing performance compared to other methods, effectively balancing fidelity to the original image and adherence to the instructions. 

\subsubsection{Evaluation on SemGen-Bench.}

To rigorously benchmark generation performance under complex semantic instructions, we evaluate different models on our proposed SemGen-Bench, which includes tasks involving long-form, complex and multi-step instructions, as shown in Table~\ref{table:4}. 
UniAlignment achieves top-2 performance across nearly all evaluated categories. 
Importantly, since our framework does not rely on external VLMs at inference, it achieves significantly faster generation time, highlighting the efficiency of its unified diffusion-based architecture.
These results further validate UniAlignment’s strength in visual-language semantic understanding and its effectiveness in handling semantically challenging multimodal tasks.

\subsubsection{Image Perception and Personalization.}

\begin{figure}[t]
    \centering
    \vspace{-6pt}
    \includegraphics[width=1.0\linewidth]{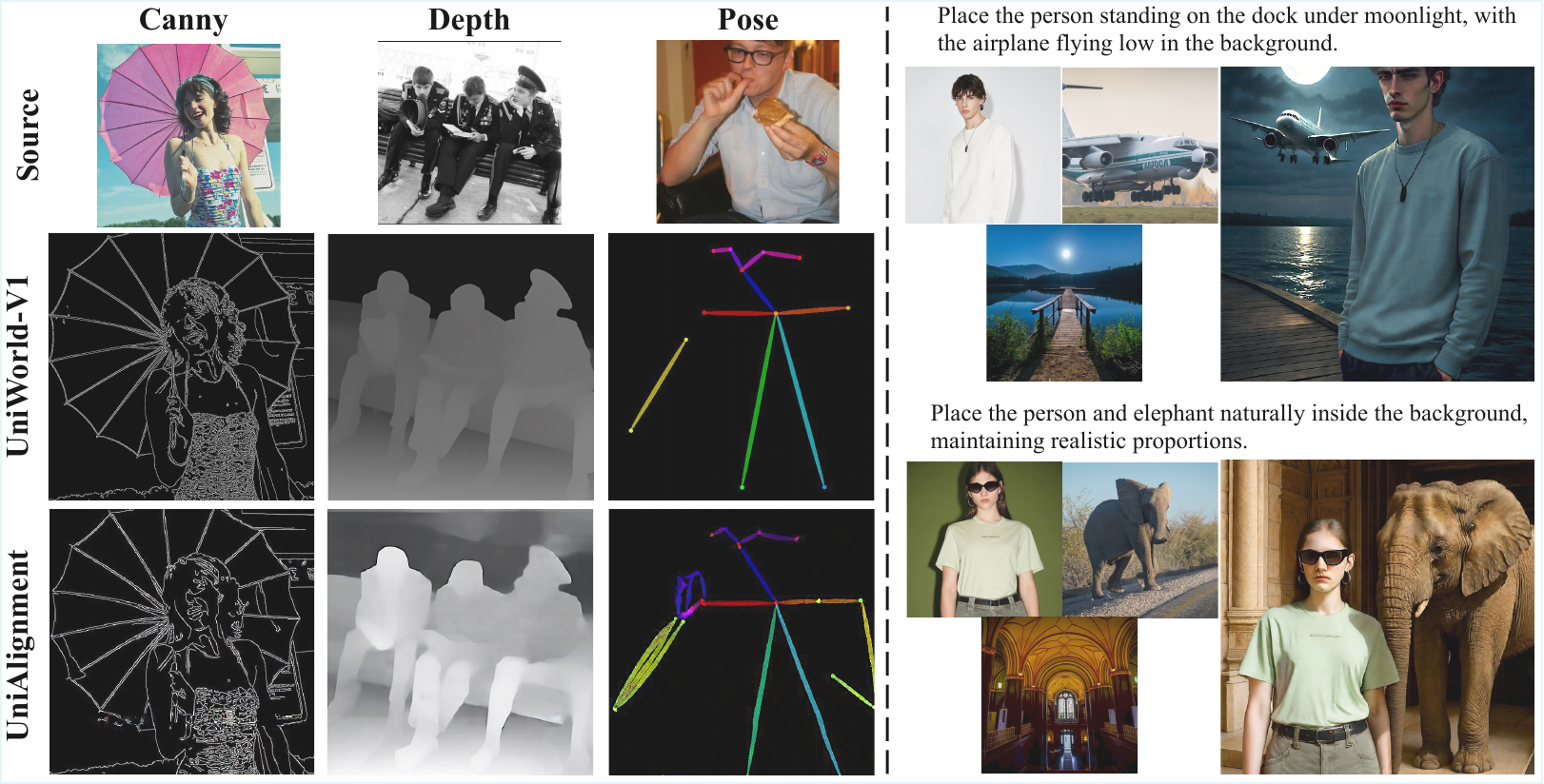}
    \vspace{-12pt}
    \caption{\textbf{Showcases of image perception and personalization.} UniAlignment is capable of predicting canny, depth, pose as well as single and multiple subject personalization. Image perception results are compared with UniWorld.}
    \vspace{-2pt}
    \label{fig:7}
\end{figure}

Besides common tasks such as text-to-image generation and image editing, UniAlignment is also capable of realizing image perception as well as subject personalization. As shown in Fig.~\ref{fig:7}, UniAlignment is capable of generating more refined and visually distinctive image-perception results compared to UniWorld, while also excelling at multi-subject customized image generation tasks.

\begin{figure}[t]
    \centering
    \includegraphics[width=1.0\linewidth]{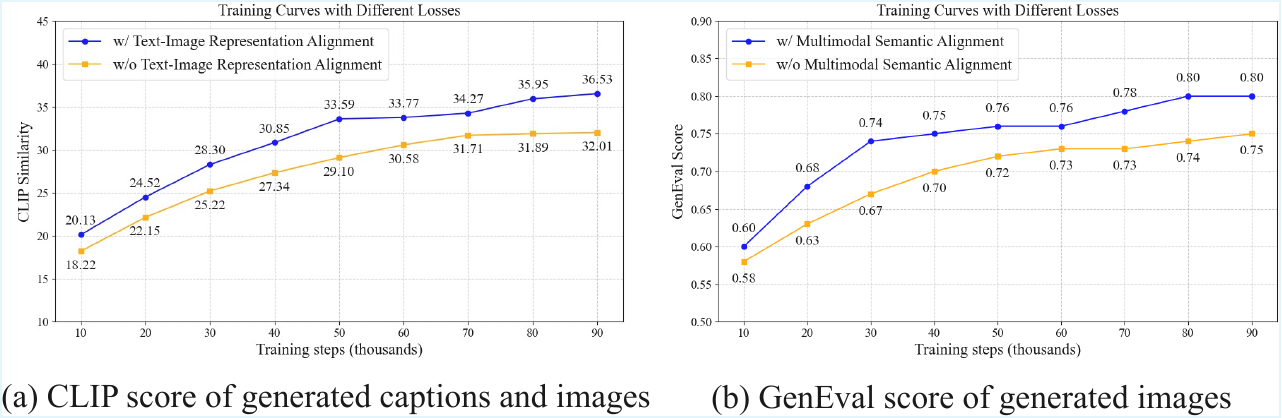}
    \vspace{-12pt}
    \caption{\textbf{Ablation results of semantic alignment.}}
    \vspace{-12pt}
    \label{fig:8}
\end{figure}

\subsection{Ablation Study}
We conduct ablation study of our proposed two semantic alignment mechanisms. As illustrated in Fig.~\ref{fig:8}, training with the cross-modal semantic alignment encourages the joint modelling of visual and textual modalities within a shared latent space, mitigating the optimization conflict and enhancing instruction consistency. Meanwhile, the intrinsic-modal semantic alignment enriches the latent semantics during the diffusion denoising process, leading to improved semantic grounding for image generation.

\section{Conclusion}

In this work, we propose UniAlignment, a unified generative model for multimodal generation. 
Built upon a single lightweight diffusion transformer, UniAlignment is capable of handling a broad spectrum of multimodal tasks, including image generation, understanding, editing, perception, and personalization.
Two semantic alignment mechanisms are introduced to enhance the semantic grounding and image-text consistency, without introducing extra parameters. 
Furthermore, a multi-stage training strategy is implemented to enable progressive task specialization. 
Experimental results across standard benchmarks as well as our proposed SemGen-Bench demonstrate the superior performance of UniAlignment, offering a promising direction toward general-purpose multimodal intelligence.

\bibliography{main}

\begin{thebibliography}{65}
\providecommand{\natexlab}[1]{#1}

\bibitem[{AI et~al.(2025)AI, Gong, Zou, Zheng, Yu, Chen, Sun, Zhao, Zhou, Ji et~al.}]{ai2025ming}
AI, I.; Gong, B.; Zou, C.; Zheng, D.; Yu, H.; Chen, J.; Sun, J.; Zhao, J.; Zhou, J.; Ji, K.; et~al. 2025.
\newblock Ming-lite-uni: Advancements in unified architecture for natural multimodal interaction.
\newblock \emph{arXiv preprint arXiv:2505.02471}.

\bibitem[{Bai et~al.(2023)Bai, Bai, Chu, Cui, Dang, Deng, Fan, Ge, Han, Huang et~al.}]{bai2023qwen}
Bai, J.; Bai, S.; Chu, Y.; Cui, Z.; Dang, K.; Deng, X.; Fan, Y.; Ge, W.; Han, Y.; Huang, F.; et~al. 2023.
\newblock Qwen technical report.
\newblock \emph{arXiv preprint arXiv:2309.16609}.

\bibitem[{Bai et~al.(2025)Bai, Chen, Liu, Wang, Ge, Song, Dang, Wang, Wang, Tang et~al.}]{bai2025qwen2}
Bai, S.; Chen, K.; Liu, X.; Wang, J.; Ge, W.; Song, S.; Dang, K.; Wang, P.; Wang, S.; Tang, J.; et~al. 2025.
\newblock Qwen2. 5-vl technical report.
\newblock \emph{arXiv preprint arXiv:2502.13923}.

\bibitem[{{Black Forest Labs}(2024)}]{Flux}
{Black Forest Labs}. 2024.
\newblock Flux.
\newblock \url{https://github.com/black-forest-labs/flux}.

\bibitem[{Brooks, Holynski, and Efros(2023)}]{brooks2023instructpix2pix}
Brooks, T.; Holynski, A.; and Efros, A.~A. 2023.
\newblock Instructpix2pix: Learning to follow image editing instructions.
\newblock In \emph{Proceedings of the IEEE/CVF conference on computer vision and pattern recognition}, 18392--18402.

\bibitem[{Caron et~al.(2021)Caron, Touvron, Misra, J{\'e}gou, Mairal, Bojanowski, and Joulin}]{caron2021emerging}
Caron, M.; Touvron, H.; Misra, I.; J{\'e}gou, H.; Mairal, J.; Bojanowski, P.; and Joulin, A. 2021.
\newblock Emerging properties in self-supervised vision transformers.
\newblock In \emph{Proceedings of the IEEE/CVF international conference on computer vision}, 9650--9660.

\bibitem[{Chen et~al.(2025{\natexlab{a}})Chen, Cai, Chen, Chen, Ji, Wang, Yang, and Wang}]{chen2025sharegpt}
Chen, J.; Cai, Z.; Chen, P.; Chen, S.; Ji, K.; Wang, X.; Yang, Y.; and Wang, B. 2025{\natexlab{a}}.
\newblock ShareGPT-4o-Image: Aligning Multimodal Models with GPT-4o-Level Image Generation.
\newblock \emph{arXiv preprint arXiv:2506.18095}.

\bibitem[{Chen et~al.(2025{\natexlab{b}})Chen, Xu, Pan, Hu, Qin, Goldstein, Huang, Zhou, Xie, Savarese et~al.}]{chen2025blip3}
Chen, J.; Xu, Z.; Pan, X.; Hu, Y.; Qin, C.; Goldstein, T.; Huang, L.; Zhou, T.; Xie, S.; Savarese, S.; et~al. 2025{\natexlab{b}}.
\newblock Blip3-o: A family of fully open unified multimodal models-architecture, training and dataset.
\newblock \emph{arXiv preprint arXiv:2505.09568}.

\bibitem[{Chen et~al.(2025{\natexlab{c}})Chen, Wu, Liu, Pan, Liu, Xie, Yu, and Ruan}]{chen2025janus}
Chen, X.; Wu, Z.; Liu, X.; Pan, Z.; Liu, W.; Xie, Z.; Yu, X.; and Ruan, C. 2025{\natexlab{c}}.
\newblock Janus-pro: Unified multimodal understanding and generation with data and model scaling.
\newblock \emph{arXiv preprint arXiv:2501.17811}.

\bibitem[{Esser et~al.(2024)Esser, Kulal, Blattmann, Entezari, M{\"u}ller, Saini, Levi, Lorenz, Sauer, Boesel et~al.}]{esser2024scaling}
Esser, P.; Kulal, S.; Blattmann, A.; Entezari, R.; M{\"u}ller, J.; Saini, H.; Levi, Y.; Lorenz, D.; Sauer, A.; Boesel, F.; et~al. 2024.
\newblock Scaling rectified flow transformers for high-resolution image synthesis.
\newblock In \emph{Forty-first international conference on machine learning}.

\bibitem[{Ghosh, Hajishirzi, and Schmidt(2023)}]{ghosh2023geneval}
Ghosh, D.; Hajishirzi, H.; and Schmidt, L. 2023.
\newblock Geneval: An object-focused framework for evaluating text-to-image alignment.
\newblock \emph{Advances in Neural Information Processing Systems}, 36: 52132--52152.

\bibitem[{{Google}(2025)}]{Gemini}
{Google}. 2025.
\newblock Gemini 2.0 flash.
\newblock \url{https://developers.googleblog.com/en/ experiment-with-gemini-20-flash-native-image-generation}.

\bibitem[{Hu et~al.(2024)Hu, Wang, Fang, Fu, Cheng, and Yu}]{hu2024ella}
Hu, X.; Wang, R.; Fang, Y.; Fu, B.; Cheng, P.; and Yu, G. 2024.
\newblock Ella: Equip diffusion models with llm for enhanced semantic alignment.
\newblock \emph{arXiv preprint arXiv:2403.05135}.

\bibitem[{Huang et~al.(2025)Huang, Wang, Yang, Lu, Yuan, Han, Hou, Zhang, Hong, Zhao et~al.}]{huang2025illume+}
Huang, R.; Wang, C.; Yang, J.; Lu, G.; Yuan, Y.; Han, J.; Hou, L.; Zhang, W.; Hong, L.; Zhao, H.; et~al. 2025.
\newblock Illume+: Illuminating unified mllm with dual visual tokenization and diffusion refinement.
\newblock \emph{arXiv preprint arXiv:2504.01934}.

\bibitem[{Hudson et~al.(2024)Hudson, Zoran, Malinowski, Lampinen, Jaegle, McClelland, Matthey, Hill, and Lerchner}]{hudson2024soda}
Hudson, D.~A.; Zoran, D.; Malinowski, M.; Lampinen, A.~K.; Jaegle, A.; McClelland, J.~L.; Matthey, L.; Hill, F.; and Lerchner, A. 2024.
\newblock Soda: Bottleneck diffusion models for representation learning.
\newblock In \emph{Proceedings of the IEEE/CVF Conference on Computer Vision and Pattern Recognition}, 23115--23127.

\bibitem[{{Jacky Hate}(2024)}]{T2I2M}
{Jacky Hate}. 2024.
\newblock Text-to-image-2m dataset.
\newblock \url{https://huggingface.co/datasets/jackyhate/text-to-image-2M}.

\bibitem[{Jia et~al.(2021)Jia, Yang, Xia, Chen, Parekh, Pham, Le, Sung, Li, and Duerig}]{jia2021scaling}
Jia, C.; Yang, Y.; Xia, Y.; Chen, Y.-T.; Parekh, Z.; Pham, H.; Le, Q.; Sung, Y.-H.; Li, Z.; and Duerig, T. 2021.
\newblock Scaling up visual and vision-language representation learning with noisy text supervision.
\newblock In \emph{International conference on machine learning}, 4904--4916. PMLR.

\bibitem[{Ku et~al.(2024)Ku, Jiang, Wei, Yue, and Chen}]{ku2024viescore}
Ku, M.; Jiang, D.; Wei, C.; Yue, X.; and Chen, W. 2024.
\newblock VIEScore: Towards Explainable Metrics for Conditional Image Synthesis Evaluation.
\newblock In \emph{Proceedings of the 62nd Annual Meeting of the Association for Computational Linguistics (Volume 1: Long Papers)}, 12268--12290.

\bibitem[{Lee et~al.(2025)Lee, Cha, Kim, and Ye}]{lee2025aligning}
Lee, J.-Y.; Cha, B.; Kim, J.; and Ye, J.~C. 2025.
\newblock Aligning text to image in diffusion models is easier than you think.
\newblock \emph{arXiv preprint arXiv:2503.08250}.

\bibitem[{Li et~al.(2024)Li, Zhang, Guo, Zhang, Li, Zhang, Zhang, Zhang, Li, Liu et~al.}]{li2024llava}
Li, B.; Zhang, Y.; Guo, D.; Zhang, R.; Li, F.; Zhang, H.; Zhang, K.; Zhang, P.; Li, Y.; Liu, Z.; et~al. 2024.
\newblock Llava-onevision: Easy visual task transfer.
\newblock \emph{arXiv preprint arXiv:2408.03326}.

\bibitem[{Li et~al.(2023)Li, Li, Savarese, and Hoi}]{li2023blip}
Li, J.; Li, D.; Savarese, S.; and Hoi, S. 2023.
\newblock Blip-2: Bootstrapping language-image pre-training with frozen image encoders and large language models.
\newblock In \emph{International conference on machine learning}, 19730--19742. PMLR.

\bibitem[{Li et~al.(2022)Li, Li, Xiong, and Hoi}]{li2022blip}
Li, J.; Li, D.; Xiong, C.; and Hoi, S. 2022.
\newblock Blip: Bootstrapping language-image pre-training for unified vision-language understanding and generation.
\newblock In \emph{International conference on machine learning}, 12888--12900. PMLR.

\bibitem[{Li et~al.(2025)Li, Li, Shi, Farimani, Kluger, Yang, and Wang}]{li2025dual}
Li, Z.; Li, H.; Shi, Y.; Farimani, A.~B.; Kluger, Y.; Yang, L.; and Wang, P. 2025.
\newblock Dual diffusion for unified image generation and understanding.
\newblock In \emph{Proceedings of the Computer Vision and Pattern Recognition Conference}, 2779--2790.

\bibitem[{Lin et~al.(2025)Lin, Li, Cheng, Niu, Ye, He, Yuan, Yu, Wang, Ge et~al.}]{lin2025uniworld}
Lin, B.; Li, Z.; Cheng, X.; Niu, Y.; Ye, Y.; He, X.; Yuan, S.; Yu, W.; Wang, S.; Ge, Y.; et~al. 2025.
\newblock Uniworld: High-resolution semantic encoders for unified visual understanding and generation.
\newblock \emph{arXiv preprint arXiv:2506.03147}.

\bibitem[{Lin et~al.(2014)Lin, Maire, Belongie, Hays, Perona, Ramanan, Doll{\'a}r, and Zitnick}]{lin2014microsoft}
Lin, T.-Y.; Maire, M.; Belongie, S.; Hays, J.; Perona, P.; Ramanan, D.; Doll{\'a}r, P.; and Zitnick, C.~L. 2014.
\newblock Microsoft coco: Common objects in context.
\newblock In \emph{European conference on computer vision}, 740--755. Springer.

\bibitem[{Lipman et~al.(2022)Lipman, Chen, Ben-Hamu, Nickel, and Le}]{lipmanflow}
Lipman, Y.; Chen, R.~T.; Ben-Hamu, H.; Nickel, M.; and Le, M. 2022.
\newblock Flow Matching for Generative Modeling.
\newblock In \emph{The Eleventh International Conference on Learning Representations}.

\bibitem[{Liu et~al.(2025)Liu, Han, Xing, Yin, Wang, Cheng, Liao, Wang, Fu, Han et~al.}]{liu2025step1x}
Liu, S.; Han, Y.; Xing, P.; Yin, F.; Wang, R.; Cheng, W.; Liao, J.; Wang, Y.; Fu, H.; Han, C.; et~al. 2025.
\newblock Step1x-edit: A practical framework for general image editing.
\newblock \emph{arXiv preprint arXiv:2504.17761}.

\bibitem[{Ma et~al.(2025)Ma, Ge, Wang, Guo, Ge, and Shan}]{ma2025genhancer}
Ma, S.; Ge, Y.; Wang, T.; Guo, Y.; Ge, Y.; and Shan, Y. 2025.
\newblock GenHancer: Imperfect generative models are secretly strong vision-centric enhancers.
\newblock \emph{arXiv preprint arXiv:2503.19480}.

\bibitem[{Mao et~al.(2025)Mao, Zhang, Pan, Jiang, Han, Liu, and Zhou}]{mao2025ace++}
Mao, C.; Zhang, J.; Pan, Y.; Jiang, Z.; Han, Z.; Liu, Y.; and Zhou, J. 2025.
\newblock Ace++: Instruction-based image creation and editing via context-aware content filling.
\newblock \emph{arXiv preprint arXiv:2501.02487}.

\bibitem[{{OpenAI}(2025)}]{Gpt4o}
{OpenAI}. 2025.
\newblock Gpt-4o.
\newblock \url{https://openai.com/index/introducing-4o-image-generation}.

\bibitem[{Pan et~al.(2025)Pan, Shukla, Singh, Zhao, Mishra, Wang, Xu, Chen, Li, Juefei-Xu et~al.}]{pan2025transfer}
Pan, X.; Shukla, S.~N.; Singh, A.; Zhao, Z.; Mishra, S.~K.; Wang, J.; Xu, Z.; Chen, J.; Li, K.; Juefei-Xu, F.; et~al. 2025.
\newblock Transfer between modalities with metaqueries.
\newblock \emph{arXiv preprint arXiv:2504.06256}.

\bibitem[{Peebles and Xie(2023)}]{peebles2023scalable}
Peebles, W.; and Xie, S. 2023.
\newblock Scalable diffusion models with transformers.
\newblock In \emph{Proceedings of the IEEE/CVF international conference on computer vision}, 4195--4205.

\bibitem[{Qu et~al.(2025)Qu, Zhang, Liu, Wang, Jiang, Gao, Ye, Du, Yuan, and Wu}]{qu2025tokenflow}
Qu, L.; Zhang, H.; Liu, Y.; Wang, X.; Jiang, Y.; Gao, Y.; Ye, H.; Du, D.~K.; Yuan, Z.; and Wu, X. 2025.
\newblock Tokenflow: Unified image tokenizer for multimodal understanding and generation.
\newblock In \emph{Proceedings of the Computer Vision and Pattern Recognition Conference}, 2545--2555.

\bibitem[{Radford et~al.(2021)Radford, Kim, Hallacy, Ramesh, Goh, Agarwal, Sastry, Askell, Mishkin, Clark et~al.}]{radford2021learning}
Radford, A.; Kim, J.~W.; Hallacy, C.; Ramesh, A.; Goh, G.; Agarwal, S.; Sastry, G.; Askell, A.; Mishkin, P.; Clark, J.; et~al. 2021.
\newblock Learning transferable visual models from natural language supervision.
\newblock In \emph{International conference on machine learning}, 8748--8763. PmLR.

\bibitem[{Rombach et~al.(2022)Rombach, Blattmann, Lorenz, Esser, and Ommer}]{rombach2022high}
Rombach, R.; Blattmann, A.; Lorenz, D.; Esser, P.; and Ommer, B. 2022.
\newblock High-resolution image synthesis with latent diffusion models.
\newblock In \emph{Proceedings of the IEEE/CVF conference on computer vision and pattern recognition}, 10684--10695.

\bibitem[{Shi, Wang, and Huang(2024)}]{shi2024seededit}
Shi, Y.; Wang, P.; and Huang, W. 2024.
\newblock Seededit: Align image re-generation to image editing.
\newblock \emph{arXiv preprint arXiv:2411.06686}.

\bibitem[{Sun et~al.(2023)Sun, Fang, Wu, Wang, and Cao}]{sun2023eva}
Sun, Q.; Fang, Y.; Wu, L.; Wang, X.; and Cao, Y. 2023.
\newblock Eva-clip: Improved training techniques for clip at scale.
\newblock \emph{arXiv preprint arXiv:2303.15389}.

\bibitem[{Swerdlow et~al.(2025)Swerdlow, Prabhudesai, Gandhi, Pathak, and Fragkiadaki}]{swerdlow2025unified}
Swerdlow, A.; Prabhudesai, M.; Gandhi, S.; Pathak, D.; and Fragkiadaki, K. 2025.
\newblock Unified multimodal discrete diffusion.
\newblock \emph{arXiv preprint arXiv:2503.20853}.

\bibitem[{Team(2024)}]{team2024chameleon}
Team, C. 2024.
\newblock Chameleon: Mixed-modal early-fusion foundation models.
\newblock \emph{arXiv preprint arXiv:2405.09818}.

\bibitem[{Tian et~al.(2024)Tian, Tao, Dai, Li, Li, Lu, Wang, Li, Huang, and Zhu}]{tian2024addp}
Tian, C.; Tao, C.; Dai, J.; Li, H.; Li, Z.; Lu, L.; Wang, X.; Li, H.; Huang, G.; and Zhu, X. 2024.
\newblock ADDP: Learning General Representations for Image Recognition and Generation with Alternating Denoising Diffusion Process.
\newblock In \emph{ICLR}.

\bibitem[{Touvron et~al.(2023)Touvron, Lavril, Izacard, Martinet, Lachaux, Lacroix, Rozi{\`e}re, Goyal, Hambro, Azhar et~al.}]{touvron2023llama}
Touvron, H.; Lavril, T.; Izacard, G.; Martinet, X.; Lachaux, M.-A.; Lacroix, T.; Rozi{\`e}re, B.; Goyal, N.; Hambro, E.; Azhar, F.; et~al. 2023.
\newblock Llama: Open and efficient foundation language models.
\newblock \emph{arXiv preprint arXiv:2302.13971}.

\bibitem[{Wang et~al.(2025{\natexlab{a}})Wang, Li, Wang, Zhao, and Zhang}]{wang2025mind}
Wang, S.; Li, W.; Wang, Q.; Zhao, S.; and Zhang, J. 2025{\natexlab{a}}.
\newblock MIND-Edit: MLLM Insight-Driven Editing via Language-Vision Projection.
\newblock \emph{arXiv preprint arXiv:2505.19149}.

\bibitem[{Wang et~al.(2025{\natexlab{b}})Wang, Sun, Zhang, Tang, Liu, and Wang}]{wangdiffusion}
Wang, W.; Sun, Q.; Zhang, F.; Tang, Y.; Liu, J.; and Wang, X. 2025{\natexlab{b}}.
\newblock Diffusion Feedback Helps CLIP See Better.
\newblock In \emph{The Thirteenth International Conference on Learning Representations}.

\bibitem[{Wang et~al.(2024)Wang, Zhang, Luo, Sun, Cui, Wang, Zhang, Wang, Li, Yu et~al.}]{wang2024emu3}
Wang, X.; Zhang, X.; Luo, Z.; Sun, Q.; Cui, Y.; Wang, J.; Zhang, F.; Wang, Y.; Li, Z.; Yu, Q.; et~al. 2024.
\newblock Emu3: Next-token prediction is all you need.
\newblock \emph{arXiv preprint arXiv:2409.18869}.

\bibitem[{Wei et~al.(2023)Wei, Mangalam, Huang, Li, Fan, Xu, Wang, Xie, Yuille, and Feichtenhofer}]{wei2023diffusion}
Wei, C.; Mangalam, K.; Huang, P.-Y.; Li, Y.; Fan, H.; Xu, H.; Wang, H.; Xie, C.; Yuille, A.; and Feichtenhofer, C. 2023.
\newblock Diffusion models as masked autoencoders.
\newblock In \emph{Proceedings of the IEEE/CVF International Conference on Computer Vision}, 16284--16294.

\bibitem[{Wei et~al.(2024)Wei, Xiong, Ren, Du, Zhang, and Chen}]{wei2024omniedit}
Wei, C.; Xiong, Z.; Ren, W.; Du, X.; Zhang, G.; and Chen, W. 2024.
\newblock Omniedit: Building image editing generalist models through specialist supervision.
\newblock In \emph{The Thirteenth International Conference on Learning Representations}.

\bibitem[{Wu et~al.(2025{\natexlab{a}})Wu, Chen, Wu, Ma, Liu, Pan, Liu, Xie, Yu, Ruan et~al.}]{wu2025janus}
Wu, C.; Chen, X.; Wu, Z.; Ma, Y.; Liu, X.; Pan, Z.; Liu, W.; Xie, Z.; Yu, X.; Ruan, C.; et~al. 2025{\natexlab{a}}.
\newblock Janus: Decoupling visual encoding for unified multimodal understanding and generation.
\newblock In \emph{Proceedings of the Computer Vision and Pattern Recognition Conference}, 12966--12977.

\bibitem[{Wu et~al.(2025{\natexlab{b}})Wu, Zheng, Yan, Xiao, Luo, Wang, Li, Jiang, Liu, Zhou et~al.}]{wu2025omnigen2}
Wu, C.; Zheng, P.; Yan, R.; Xiao, S.; Luo, X.; Wang, Y.; Li, W.; Jiang, X.; Liu, Y.; Zhou, J.; et~al. 2025{\natexlab{b}}.
\newblock OmniGen2: Exploration to Advanced Multimodal Generation.
\newblock \emph{arXiv preprint arXiv:2506.18871}.

\bibitem[{Wu et~al.(2025{\natexlab{c}})Wu, Zhang, Chen, Tang, Li, Fang, Zhu, Xie, Yin, Yi et~al.}]{wu2024vila}
Wu, Y.; Zhang, Z.; Chen, J.; Tang, H.; Li, D.; Fang, Y.; Zhu, L.; Xie, E.; Yin, H.; Yi, L.; et~al. 2025{\natexlab{c}}.
\newblock Vila-u: a unified foundation model integrating visual understanding and generation.
\newblock In \emph{The Thirteenth International Conference on Learning Representations}.

\bibitem[{Xiang et~al.(2023)Xiang, Yang, Huang, and Wang}]{xiang2023denoising}
Xiang, W.; Yang, H.; Huang, D.; and Wang, Y. 2023.
\newblock Denoising diffusion autoencoders are unified self-supervised learners.
\newblock In \emph{Proceedings of the IEEE/CVF International Conference on Computer Vision}, 15802--15812.

\bibitem[{Xiao et~al.(2025)Xiao, Wang, Zhou, Yuan, Xing, Yan, Li, Wang, Huang, and Liu}]{xiao2025omnigen}
Xiao, S.; Wang, Y.; Zhou, J.; Yuan, H.; Xing, X.; Yan, R.; Li, C.; Wang, S.; Huang, T.; and Liu, Z. 2025.
\newblock Omnigen: Unified image generation.
\newblock In \emph{Proceedings of the Computer Vision and Pattern Recognition Conference}, 13294--13304.

\bibitem[{Xie et~al.(2024{\natexlab{a}})Xie, Mao, Bai, Zhang, Wang, Lin, Gu, Chen, Yang, and Shou}]{xie2024show}
Xie, J.; Mao, W.; Bai, Z.; Zhang, D.~J.; Wang, W.; Lin, K.~Q.; Gu, Y.; Chen, Z.; Yang, Z.; and Shou, M.~Z. 2024{\natexlab{a}}.
\newblock Show-o: One single transformer to unify multimodal understanding and generation.
\newblock \emph{arXiv preprint arXiv:2408.12528}.

\bibitem[{Xie et~al.(2024{\natexlab{b}})Xie, Du, Song, and Liu}]{xie2024muse}
Xie, R.; Du, C.; Song, P.; and Liu, C. 2024{\natexlab{b}}.
\newblock Muse-vl: Modeling unified vlm through semantic discrete encoding.
\newblock \emph{arXiv preprint arXiv:2411.17762}.

\bibitem[{Yang et~al.(2025)Yang, Tian, Li, Zhang, Shen, Tong, and Wang}]{yang2025mmada}
Yang, L.; Tian, Y.; Li, B.; Zhang, X.; Shen, K.; Tong, Y.; and Wang, M. 2025.
\newblock Mmada: Multimodal large diffusion language models.
\newblock \emph{arXiv preprint arXiv:2505.15809}.

\bibitem[{Yao, Yang, and Wang(2025)}]{yao2025reconstruction}
Yao, J.; Yang, B.; and Wang, X. 2025.
\newblock Reconstruction vs. generation: Taming optimization dilemma in latent diffusion models.
\newblock In \emph{Proceedings of the Computer Vision and Pattern Recognition Conference}, 15703--15712.

\bibitem[{Ye et~al.(2025)Ye, He, Li, Lin, Yuan, Yan, Hou, and Yuan}]{ye2025imgedit}
Ye, Y.; He, X.; Li, Z.; Lin, B.; Yuan, S.; Yan, Z.; Hou, B.; and Yuan, L. 2025.
\newblock Imgedit: A unified image editing dataset and benchmark.
\newblock \emph{arXiv preprint arXiv:2505.20275}.

\bibitem[{Yu et~al.(2025{\natexlab{a}})Yu, Chow, Yue, Pan, Wu, Wan, Li, Tang, Zhang, and Zhuang}]{yu2025anyedit}
Yu, Q.; Chow, W.; Yue, Z.; Pan, K.; Wu, Y.; Wan, X.; Li, J.; Tang, S.; Zhang, H.; and Zhuang, Y. 2025{\natexlab{a}}.
\newblock Anyedit: Mastering unified high-quality image editing for any idea.
\newblock In \emph{Proceedings of the Computer Vision and Pattern Recognition Conference}, 26125--26135.

\bibitem[{Yu et~al.(2025{\natexlab{b}})Yu, Kwak, Jang, Jeong, Huang, Shin, and Xie}]{yurepresentation}
Yu, S.; Kwak, S.; Jang, H.; Jeong, J.; Huang, J.; Shin, J.; and Xie, S. 2025{\natexlab{b}}.
\newblock Representation Alignment for Generation: Training Diffusion Transformers Is Easier Than You Think.
\newblock In \emph{The Thirteenth International Conference on Learning Representations}.

\bibitem[{Yu et~al.(2024)Yu, Zeng, Hua, Fu, and Luo}]{yu2024promptfix}
Yu, Y.; Zeng, Z.; Hua, H.; Fu, J.; and Luo, J. 2024.
\newblock PromptFix: you prompt and we fix the photo.
\newblock In \emph{Proceedings of the 38th International Conference on Neural Information Processing Systems}, 40000--40031.

\bibitem[{Zhai et~al.(2023)Zhai, Mustafa, Kolesnikov, and Beyer}]{zhai2023sigmoid}
Zhai, X.; Mustafa, B.; Kolesnikov, A.; and Beyer, L. 2023.
\newblock Sigmoid loss for language image pre-training.
\newblock In \emph{Proceedings of the IEEE/CVF international conference on computer vision}, 11975--11986.

\bibitem[{Zhang et~al.(2025)Zhang, Duan, Wang, Zhao, Lu, Di, Xu, Chen, and Zhang}]{zhang2025nexus}
Zhang, H.; Duan, Z.; Wang, X.; Zhao, Y.; Lu, W.; Di, Z.; Xu, Y.; Chen, Y.; and Zhang, Y. 2025.
\newblock Nexus-gen: A unified model for image understanding, generation, and editing.
\newblock \emph{arXiv preprint arXiv:2504.21356}.

\bibitem[{Zhang et~al.(2023)Zhang, Mo, Chen, Sun, and Su}]{zhang2023magicbrush}
Zhang, K.; Mo, L.; Chen, W.; Sun, H.; and Su, Y. 2023.
\newblock Magicbrush: A manually annotated dataset for instruction-guided image editing.
\newblock \emph{Advances in Neural Information Processing Systems}, 36: 31428--31449.

\bibitem[{Zhang et~al.(2024{\natexlab{a}})Zhang, Zhang, Xu, and Tao}]{zhang2024vision}
Zhang, Q.; Zhang, J.; Xu, Y.; and Tao, D. 2024{\natexlab{a}}.
\newblock Vision transformer with quadrangle attention.
\newblock \emph{IEEE Transactions on Pattern Analysis and Machine Intelligence}, 46(5): 3608--3624.

\bibitem[{Zhang et~al.(2024{\natexlab{b}})Zhang, Zhang, Xie, Li, Dai, Long, Xie, Zhang, Li, and Zhang}]{zhang2024gme}
Zhang, X.; Zhang, Y.; Xie, W.; Li, M.; Dai, Z.; Long, D.; Xie, P.; Zhang, M.; Li, W.; and Zhang, M. 2024{\natexlab{b}}.
\newblock GME: Improving Universal Multimodal Retrieval by Multimodal LLMs.
\newblock \emph{arXiv preprint arXiv:2412.16855}.

\bibitem[{Zhou et~al.(2024)Zhou, Yu, Babu, Tirumala, Yasunaga, Shamis, Kahn, Ma, Zettlemoyer, and Levy}]{zhou2024transfusion}
Zhou, C.; Yu, L.; Babu, A.; Tirumala, K.; Yasunaga, M.; Shamis, L.; Kahn, J.; Ma, X.; Zettlemoyer, L.; and Levy, O. 2024.
\newblock Transfusion: Predict the next token and diffuse images with one multi-modal model.
\newblock \emph{arXiv preprint arXiv:2408.11039}.

\end{thebibliography}

\end{document}